\documentclass[runningheads]{llncs}

\usepackage[T1]{fontenc}
\usepackage{times}
\usepackage{latexsym}
\usepackage{graphicx}
\usepackage{stmaryrd}
\usepackage{placeins}
\usepackage{silence}
\usepackage[compatibility=false]{caption}
\usepackage{subcaption}
\usepackage{hyperref}
\usepackage{multirow}
\usepackage{booktabs}
\usepackage{cancel}
\usepackage{epstopdf} 
\usepackage{xcolor}
\usepackage{silence}
\usepackage{microtype}
\usepackage{hyperref}
\newcommand{\orcid}[1]{\href{https://orcid.org/#1}{\textsuperscript{\includegraphics[height=2.5ex]{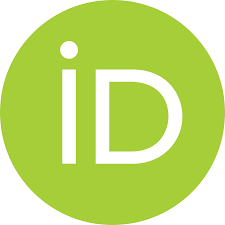}}}}

\usepackage{inconsolata}

\usepackage{amssymb}
\usepackage{amssymb, pifont}

\usepackage{color}

\begin{document}

\title{Influential Training Data Retrieval for Explaining Verbalized Confidence of LLMs}

\author{Yuxi Xia\inst{1,2}\orcid{0000-0001-5879-7209} \and
 Loris Schoenegger\inst{1,2}\orcid{0009-0003-4546-9152} \and
Benjamin Roth\inst{1,3}\orcid{0000-0003-0362-0267}}

\authorrunning{Y. Xia et al.}

\institute{Faculty of Computer Science, University of Vienna, Vienna, Austria\and
UniVie Doctoral School Computer Science, Vienna, Austria\and
Faculty of Philological and Cultural Studies, University of Vienna, Vienna, Austria\\
\texttt{\{yuxi.xia,loris.schoenegger,benjamin.roth\}@univie.ac.at}
}
\maketitle              

\begin{abstract}
Large language models (LLMs) can increase users’ perceived trust by verbalizing confidence in their outputs. 
However, prior work has shown that LLMs are often overconfident, making their stated confidence unreliable since it does not consistently align with factual accuracy.
To better understand the sources of this verbalized confidence, we introduce TracVC (\textbf{Trac}ing \textbf{V}erbalized \textbf{C}onfidence), a method that builds on information retrieval and influence estimation to trace generated confidence expressions back to the training data.
We evaluate TracVC on OLMo and Llama models in a question answering setting, proposing a new metric, content groundness, which measures the extent to which an LLM grounds its confidence in content-related training examples (relevant to the question and answer) versus in generic examples of confidence verbalization.
Our analysis reveals that OLMo2-13B is frequently influenced by confidence-related data that is lexically unrelated to the query, suggesting that it may mimic superficial linguistic expressions of certainty rather than rely on genuine content grounding.
These findings point to a fundamental limitation in current training regimes: LLMs may learn how to sound confident without learning when confidence is justified.
Our analysis provides a foundation for improving LLMs' trustworthiness in expressing more reliable confidence. 
\footnote{Source code is available at  
\url{https://github.com/Yuuxii/training_data_confidence/}.}

\end{abstract}

\section{Introduction}

\textit{Verbalized confidence} in large language models (LLMs) is increasingly used to estimate the certainty of their generated outputs to improve transparency and user trust \cite{chen-mueller-2024-quantifying,kadavath2023prompting,tian2023just}. However, recent studies have shown that LLMs frequently exhibit overconfidence, which is not aligned with factual accuracy,
resulting in the poor reliability of their expressed confidence \cite{10.1007/978-981-96-1710-4_10,xia-etal-2025-influences,xiong2024llms,zhou-etal-2024-relying}. This finding leads to a foundational question: what drives confidence in LLMs, and do LLMs understand the intended meaning of expressing confidence?

\begin{figure}[ht] 
    \centering
    \includegraphics[trim=0 0.2cm 0 0.1cm,clip,width=1\linewidth, keepaspectratio]{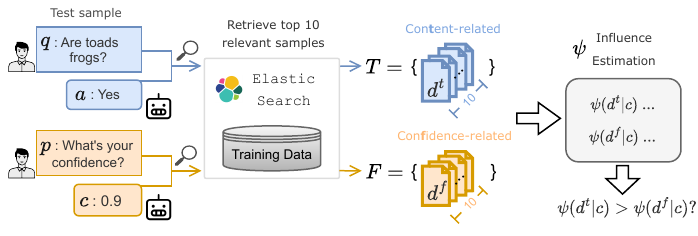}
    \caption{The workflow of TracVC. We first retrieve the top 10 relevant samples for content (question ($q$) and answer ($a$)) and confidence (prompt ($p$) and confidence ($c$)) segment in the test sample. Then, we compute and compare the influence score regarding the confidence generation for content-related and confidence-related data samples.}
    \label{fig:exp}
\end{figure}
In this paper, we investigate the origins of verbalized confidence in LLMs by examining the role of training data. Specifically, we ask: \textbf{Do LLMs ground their confidence in lexically relevant, content-related training examples, or are they more influenced by superficial, confidence-related cues?}

To address this question, we propose \textbf{TracVC} (shown in Fig. \ref{fig:exp}), a method for tracing verbalized confidence back to its influential training data. 
For each test instance, which consists of a content segment (question and LLM answer) and a verbalized confidence segment, TracVC retrieves two sets of top 10 relevant training examples: one that is lexically similar to the content and another that is similar to the confidence expression.
Then, TracVC estimates the influence of the retrieved examples on the model's confidence generation by adapting the gradient-based attribution method \textit{TracIn} \cite{pruthi_estimating_2020}.
After comparing the influence of these two sets, TracVC reveals whether content-related or confidence-related training data plays a greater role in shaping the model’s verbalized confidence.

We apply TracVC to 11 open-source LLMs with publicly available training corpora, including OLMo \cite{olmo2024furious}, Llama \cite{meta2024llama3} instruction models, and their corresponding checkpoints trained with different post-training techniques (e.g., direct preference optimization \cite{NEURIPS2023_a85b405e}). Those models are evaluated on five question answering benchmarks. To quantify the extent to which content-related training data outweighs confidence-related data in shaping verbalized confidence, we introduce \textbf{content groundness} as a new metric, defined as the proportion of cases where content-related examples dominate over confidence-related ones.
Our results demonstrate that:
\begin{enumerate}
    \item LLMs, especially the OLMo2-13B model, can be influenced by confidence-related training examples that are lexically unrelated to the question, often latching onto keywords like ``confidence'' regardless of context. This suggests that they do not fully ground their confidence in content-related information, but instead learn to mimic linguistic markers of certainty. 
    \item Larger LLMs do not demonstrate higher content groundness than smaller ones. We hypothesize that larger models, due to their higher capacity, may be more sensitive to stylistic or superficial patterns in training data.
    \item Content groundness is higher when tested on samples that are all correctly answered by the evaluated LLM. We hypothesize that LLMs are likely to ground more on content-related examples if they have seen those examples during training. 
    \item Post-training techniques can impact content groundness in the opposite direction on different LLMs.
\end{enumerate}

These findings highlight a fundamental limitation in current training regimes: \textbf{LLMs may learn to sound confident without understanding when confidence is warranted.} 
Our work introduces a data-driven perspective of model confidence, offering insights that can guide future training approaches toward improving the trustworthiness of model confidence.

\section{Related Work} 

\paragraph{Trustworthiness of Verbalized Confidence in LLMs.} Verbalized confidence offers a convenient way for users to gauge the trustworthiness of LLM outputs \cite{geng-etal-2024-survey,tian2023just,xiong2024llms}. Empirical evidence suggests that moderate confidence expressions can foster higher user trust, satisfaction, and task performance than either high or low confidence levels \cite{XU2025103455}.
However, recent studies show that LLMs often exhibit overconfidence in their answers \cite{calib-guo,xia-etal-2025-influences,xiong2024llms,zhou-etal-2024-relying}, a mismatch between expressed certainty and factual correctness that can undermine user trust. Explaining verbalized confidence is therefore crucial. 

\paragraph{Retrieving Influential Training Data for Explanation.}
Gradient-based influence estimation methods
\cite{bejan-etal-2023-make,choe2024dataworthgptllmscale,DBLP:journals/corr/abs-2205-12600,basu_influence_2020,koh_understanding_2017} provide an alternative to leave-one-out retraining for studying the effect of individual training examples on model behavior.
Those methods can broadly be grouped into two categories: (1) First-order approaches, such as \textit{TracIn} \cite{pruthi_estimating_2020}, which measure gradient similarity between training examples and test predictions; and (2) Hessian-based approaches, such as DataInf \cite{kwon_datainf_2023}, are potentially more precise, however, while 
several adaptations make such methods more scalable \cite{chang_scalable_2024,guo-etal-2021-fastif,Schioppa_Zablotskaia_Vilar_Sokolov_2022}, they are still computationally prohibitive for large-scale LLMs. 
To address scalability, information retrieval-based influence estimation methods \cite{DBLP:conf/iclr/ElazarBMRSSWGS024,liu-etal-2025-olmotrace}, such as OLMoTRACE \cite{liu-etal-2025-olmotrace}, retrieve lexically similar training examples under the assumption that they are likely to influence predictions. While efficient, such methods rely on surface-level token overlap and do not leverage internal model states as gradient-based influence estimation methods. Other approaches \cite{Chuang2025SelfCite,sourceaware2024} train models explicitly to report their influential data, but these require additional training and are unsuitable for our setting. 

Our proposed method, TracVC, utilizes both information retrieval techniques \cite{DBLP:conf/iclr/ElazarBMRSSWGS024} and the gradient-similarity-based method \textit{TracIn} \cite{pruthi_estimating_2020} to analyze how training data influences LLMs’ confidence expression, which combines the strengths of both methods and thus is efficient to apply while considering internal model states.

\paragraph{Explaining Verbalized Confidence of LLMs.}  Existing work has primarily examined verbalized confidence through the lens of its alignment with internal model probabilities \cite{kumar2024confidence,10.1007/978-981-96-1710-4_10}  or with human confidence \cite{Steyvers2024WhatLL}; others \cite{xiong2024llms,yang2025on,yin-etal-2023-large} investigate prompting strategies or calibration techniques aimed at improving the reliability of the expressed confidence.  While these approaches provide useful insights into the surface behavior of LLMs, they do not address a deeper question: what role does the training data play in shaping verbalized confidence? Our work fills this gap by tracing verbalized confidence back to influential training samples, thereby offering a data-centric perspective on the sources of LLM confidence.

\section{Methodology}

The workflow of our proposed TracVC is shown in Fig. \ref{fig:exp}. TracVC identifies which types of training data, i.e., content-related or confidence-related data, are more influential for verbalized confidence in LLMs. 
The following sections explain our prompt design, the information retrieval mechanism to retrieve related data for content and confidence from the training corpus, and finally, the method we use to estimate the influence score of the retrieved data.
\subsection{Prompt Design}
We use a two-stage prompt inspired by Tian et.al. \cite{tian2023just} for obtaining verbalized confidence. The first-stage prompt contains the question $q$ for an LLM $\mathcal{M}$ to generate ($\sim$) the answer $a$ ($a \sim \mathcal{M}(q)$). The specific prompt is \textit{``Answer the question, give ONLY the answer, no other words or explanation: $<q>$''}. The second-stage prompt consists of a confidence prompt $p$ to require $\mathcal{M}$ to provide a probability for its generated answer. The specific prompt $p$ is \textit{``Provide the probability that your answer is correct. Give ONLY the probability between 0.0 and 1.0, no other words or explanation''}. Finally, $\mathcal{M}$ generates the verbalized confidence $c$ regarding content $a$ and $q$, formulated as:
\begin{equation}
    c \sim \mathcal{M}( q, a, p).
\end{equation}
\subsection{Relevant Training Data Retrieval}
We want to verify whether \textit{content}-related training data that contains information of question and answer is more influential than \textit{confidence}-related training data that mostly contains superficial confidence cues. 
For a given test instance, we retrieve two sets of training examples: a set of con\underline{\textbf{t}}ent-related samples $T = \{d^t_1, ..., d^t_{10}\}$,
based on lexical similarity to the question $q$ and answer $a$, and a set of con\underline{\textbf{f}}idence-related samples  $F=\{d^f_1, ..., d^f_{10}\}$ based on similarity to the confidence prompt $p$ and verbalized confidence $c$ (samples are demonstrated in Fig. \ref{fig:data-examples}).

Specifically, we analyze pre-training data (\textbf{Pre}) and post-training data (\textbf{Post}) separately, as they differ in format and serve distinct training objectives. For the pre-training corpus, we retrieved examples using the inverted index over model pre-training data provided by Elazar et al.~\cite{DBLP:conf/iclr/ElazarBMRSSWGS024}. For the post-training corpora, we constructed our own inverted index with an Elasticsearch cluster, following the setup of Elazar et al.~\cite{DBLP:conf/iclr/ElazarBMRSSWGS024}. To identify relevant examples, we issued multi-match queries in \textit{best fields mode} across all indexed fields. Ranking was performed with Elasticsearch’s default scoring function, BM25 \cite{elasticsearch8172,Robertson1994OkapiAT}, using default parameters ($k_1 = 1.2$, $b = 0.75$). For each query, we retrieved the top 10 training examples ranked by relevance score.

\subsection{Training Data Influence Estimation}

Having obtained the content-related and confidence-related sets of examples, we now introduce the method to test which set more strongly influences a language model's confidence generation.
Following TracIn \cite{pruthi_estimating_2020}, our method performs efficient point-wise comparisons of loss gradients with respect to the model parameters ($\nabla \ell(w, \cdot)$), enabling our analysis to scale to large pre-training corpora. Specifically, we estimate the influence of a training sample \( d \) on the model's generated confidence \( c \), denoted as $\psi(d|c)$,  by computing the cosine similarity between loss gradients. One gradient is obtained by evaluating the loss on the training sample \( d \), while the other is computed by evaluating the loss on the completion \( (q, a, p, c ) \). The formula to get $\psi(d|c)$ is: 

\begin{equation}
\resizebox{0.6\linewidth}{!}{$
 \psi(d|c_i) = \frac{\nabla \ell(w,d) \cdot \nabla \ell(w,(q_i, a_i, p, c_i))}{\|\nabla \ell(w,d)\|\cdot \|\nabla \ell(w,(q_i, a_i, p, c_i))\|} $}.
\end{equation}

Note that whereas TracIn utilizes dot product as a similarity measure, we use cosine similarity following previous work \cite{hammoudeh_identifying_2022,hammoudeh_training_2024,park_trak_2023,xia_less_2024} to reduce the impact of gradient magnitude on our score, and compute the gradient with respect to the model's input embeddings \( w \) \cite{yeh_first_2022}. It should be emphasized that the influence score captures information beyond the word embedding layer, as the gradient chains through higher layers as well \cite{yeh_first_2022}.
\paragraph{Influence Estimation without Verbalized Confidence.} To validate our score's robustness to variations in the model’s self-reported confidence, we also report results for experiments where we omit the verbalized confidence $c$ from the completion \( (q, a, p ) \).
This computed influence score is then defined as:
\begin{equation}\label{eq:negvc}
\resizebox{0.6\linewidth}{!}{$
    \psi^{\neg c}(d|c_i) = \frac{\nabla \ell(w,d) \cdot \nabla \ell(w,(q_i, a_i, p))}{\|\nabla \ell(w,d)\|\cdot \|\nabla \ell(w,(q_i, a_i, p)\|} $}.
\end{equation}

\subsection{Experimental Setup}
Our experiments require knowledge about the training data of LLMs. Therefore, we mainly study OLMo models \cite{Groeneveld2023OLMo,olmo2024furious} that provide publicly available pre- and post-training data. Those OLMo models span different model sizes and released versions: OLMo2-13B INS (\texttt{OLMo-2-1124-13B-Instruct}), OLMo2-7B INS (\texttt{OLMo-2-1124-7B-Instruct}), OLMo-7B INS(\texttt{OLMo-7B-Instruct-hf}).  Additionally, we include Llama-3.1 \cite{meta2024llama3} ( \texttt{Llama-3.1-Tulu-3-8B} \cite{lambert2025tulu3pushingfrontiers}), which was post-trained with publicly available data.
For each model, we also evaluate its checkpoints after post-training with supervised-fine-tuning (SFT), direct preference optimization (DPO) \cite{NEURIPS2023_a85b405e}, and reinforcement learning with verified reward (RLVR) \cite{olmo2024furious} (the last training step for the instruction model). 
Our studied benchmark datasets are Natural Question (NQ) \cite{kwiatkowski2019natural}, SicQ \cite{welbl-etal-2017-crowdsourcing}, TriviaQA \cite{joshi2017triviaqa}, PopQA \cite{mallen-etal-2023-trust} and TruthfulQA \cite{lin-etal-2022-truthfulqa}. Each of the first four datasets contains 1,000 randomly selected samples, and TruthfulQA has 817.
\begin{figure}[htbp!]
    \centering
    \includegraphics[trim=0.5 0.5cm 0 0,clip,width=1\linewidth]{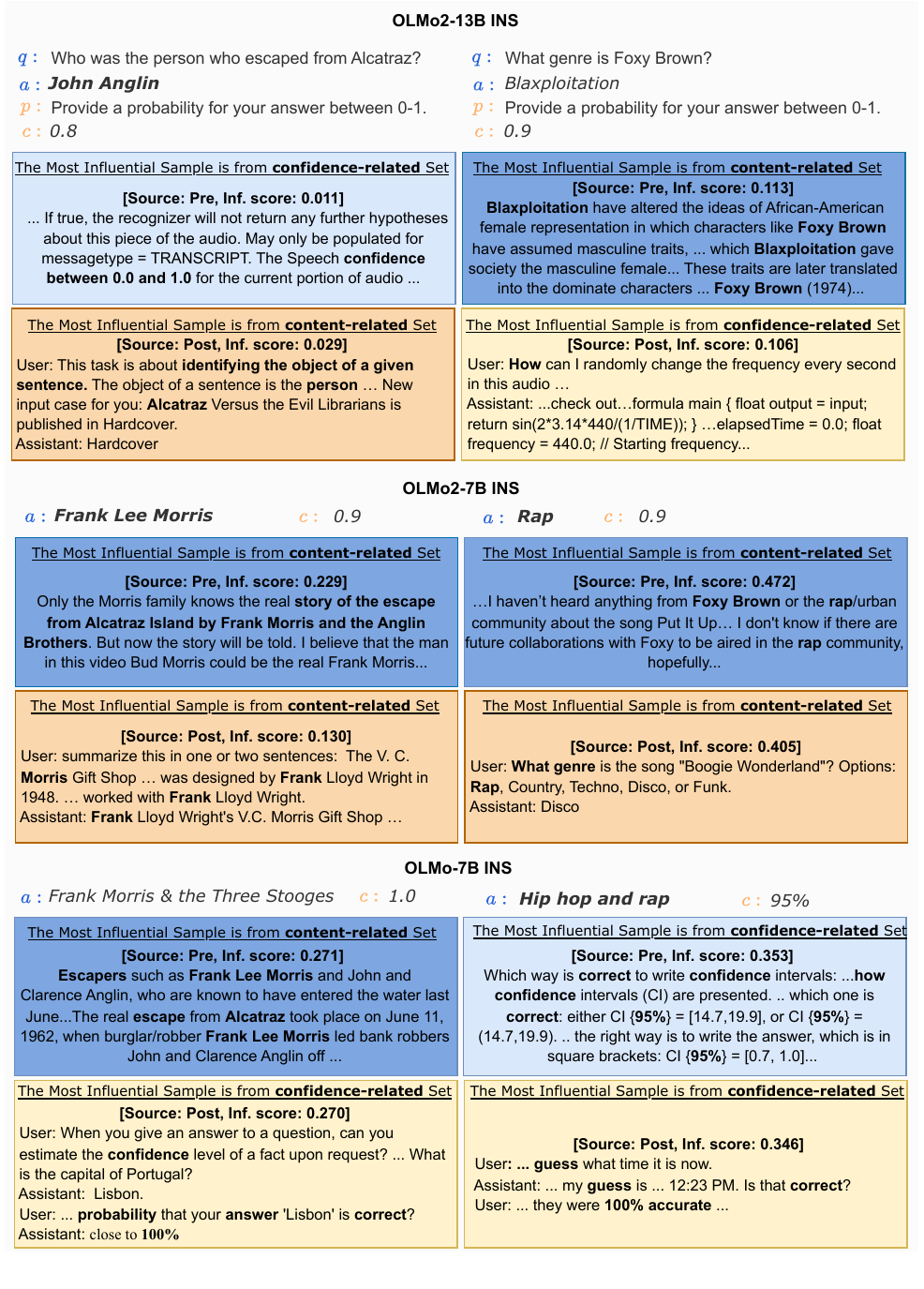}
    \caption{Examples of the most influential training samples for different LLMs when generating confidence. Retrieved samples come from pre-training (Source: Pre) and post-training (Source: Post) corpora. Ground-truth answers are shown in bold.}
    \label{fig:data-examples}
\end{figure}
We employ the vLLM \cite{vllm2023} library 
for LLM inference and serving. All LLMs are set with a temperature equal to 0 to ensure consistent outputs. All our experiments are conducted on NVIDIA HGX H100, requiring a total of approximately 642 GPU hours for testing various scenarios.

\section{Influence Analysis of Retrieved Training Data  }
We analyze the influence of retrieved data examples on model confidence predictions by presenting (1) a case study of the most influential training documents retrieved from pre-training and post-training data, (2) an overview of the proportion of the source training data for the most influential retrieved data examples, and (3) an overall statistical summary of influence scores.  

\paragraph{Case Study of the Most Influential Training Data.}
Fig.~\ref{fig:data-examples} shows the training examples with the highest estimated influence scores for two test samples. We observe that different LLMs rely on different types of examples when generating confidence estimates.
For the first example (\texttt{``Who was the person who escaped from Alcatraz?"}), all models are most influenced by content-related documents. However, for OLMo2-13B-INS, the top post-training example is irrelevant to the question or task, yielding a low influence score (0.029). By contrast, the most influential examples for the other two OLMo models come from pre-training data and are highly relevant, with influence scores exceeding 0.2. In contrast, for the second test example (\texttt{``What genre is Foxy Brown?''}), confidence-related examples get an influence score higher than 0.3 for OLMo-7B INS. This case study demonstrates that \textbf{LLMs can be affected by content-unrelated training examples that contain superficial cues, such as the keyword ``confidence''.}
\begin{figure}[ht]
    \centering
        \includegraphics[trim=0 1.0cm 0 0,clip,width=0.8\linewidth]{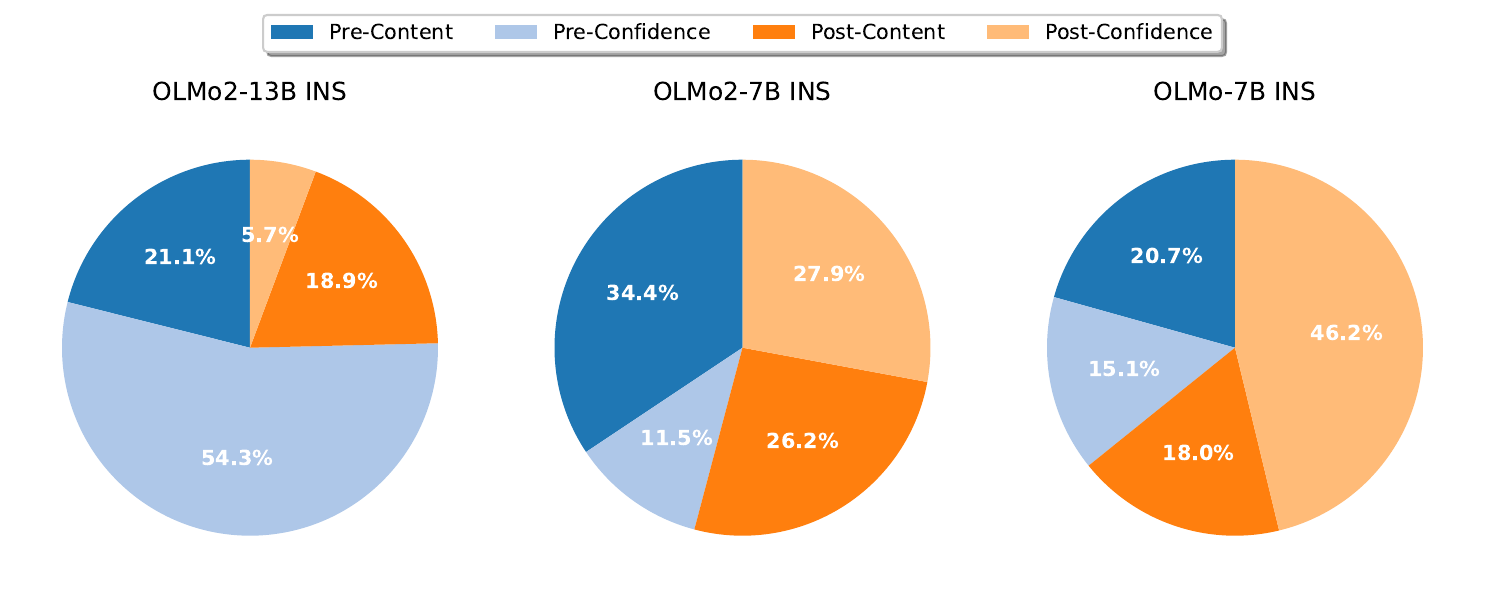}
    \caption{Proportion of sources of the most influential training examples for each test sample. The most influential example is defined as the one with the highest influence score among all retrieved examples from both pre-training and post-training corpora. For example, Pre-Content denotes a content-related example from the pre-training corpus.}
    \label{fig:win-pro}
\end{figure}
\begin{figure}[ht!]
    \centering
    \begin{subfigure}[b]{1\textwidth}
    \includegraphics[trim=0.25 0.25cm 0 0,clip,width=1\textwidth]{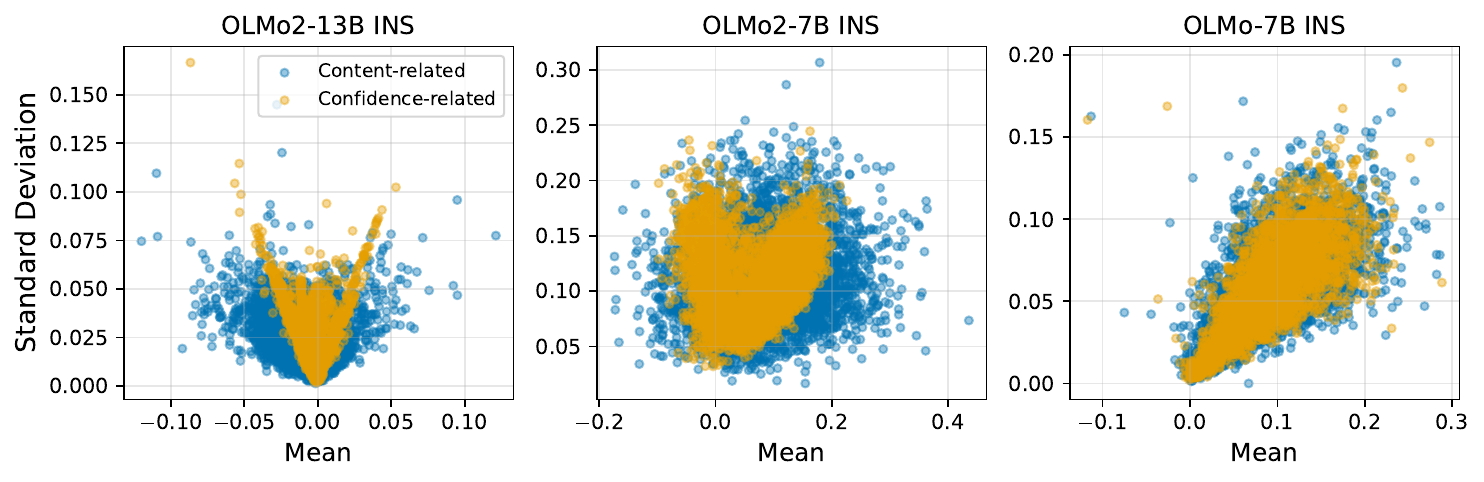}
    \caption{Pre-training data.}
    \end{subfigure}
    \begin{subfigure}[b]{1\textwidth}
    \includegraphics[trim=0.25 0.25cm 0 0,clip,width=1\textwidth]{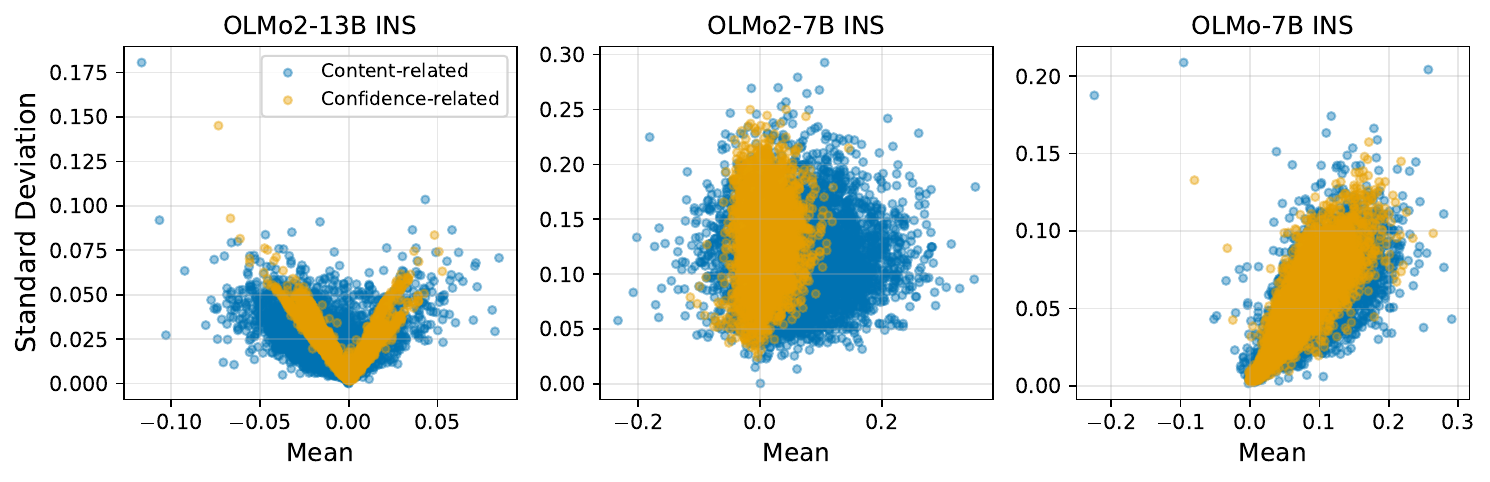}
    \caption{Post-training data.}
    \end{subfigure}
    \caption{Distribution of mean and standard deviation of influence scores across the 10 retrieved training examples for each test sample. Each point presents a test sample.}
    \label{fig:results_stat}
\end{figure}

\paragraph{Source Proportion of the Most Influential Examples.} Fig. \ref{fig:win-pro} shows that more than half of the most influential data for OLMo2-13B and OLMo-7B comes from the confidence-related example of pre-training data, while for OLMo2-7B, the biggest portion is from content-related data. We also observe that post-training is more influential for OLMo-7B and OLMo2-7B, but an opposite trend applies to OLMo2-13B.  

\paragraph{Overall Influence Statistics.}
Fig.~\ref{fig:results_stat} shows the distribution of influence scores across models. Overall, we find similar patterns between pre-training and post-training data.
For OLMo2-13B, the mean influence scores of the top 10 content-related examples are higher than those of confidence-related examples, which contradicts the earlier study based on the proportion of the most influential example. OLMo-7B shows substantial overlap between content- and confidence-related scores, while OLMo2-7B contains a larger proportion of content-related examples with higher mean scores. Notably, OLMo2-13B-INS differs from the other models by exhibiting a sizable proportion of content-related examples with relatively low mean influence scores.
These results suggest that \textbf{relying solely on the most influential examples can be misleading.} Examining aggregate statistics across the top retrieved examples provides a more comprehensive view of how LLMs ground their confidence in content- versus confidence-related training data. The following sections aim to achieve this goal.

\section{Explaining Verbalized Confidence of LLMs with Content Groundness}
Although the previous sections provide detailed case studies and statistical analyses of influence score distributions of the retrieved data, they do not fully examine how to assess the extent to which LLMs ground their verbalized confidence in content-related documents, nor the factors that impact such grounding. To fill this gap, the following section introduces a new metric for quantifying content groundness and analyzes its contributing factors, thereby offering deeper insights into the mechanisms underlying verbalized confidence.

\subsection{Measurement of Content Groundness}
We define content groundness as the property whereby content-related training data exerts a greater influence than confidence-related data on confidence generation.
To quantify groundness, we introduce \textbf{ccr} (Content-over-Confidence Ratio), defined as the winning counts ratio between the con\textbf{\underline{t}}ent-related set $T$ and con\textbf{\underline{f}}idence-related set $F$. Intuitively, a higher \textbf{ccr} indicates that the model's confidence is more strongly driven by lexically relevant content rather than by potentially misleading confidence cues in the prompt. Assume our test data contains $n$ questions, \textbf{ccr} is computed by:
\begin{equation}
\resizebox{0.6\linewidth}{!}{$
    ccr=\frac{\sum_{i=0}^n \sum_{d^t\in T_i, d^f\in F_i}\mathbf{1}(\psi(d^t|c_i)>\psi(d^f|c_i))}{\sum_{i=0}^n \sum_{d^t\in T_i, d^f\in F_i}\mathbf{1}(\psi(d^t|c_i)<\psi(d^f|c_i))}$}.
\end{equation}
We define our score as a ratio to capture relative rather than absolute importance, and use it for analysis at the aggregate level, rather than to assess individual instances.
We take this approach because individual point-wise influence measurements, while inexpensive, can be noisy and have been shown to sometimes poorly predict re-training effects for deep models not trained to convergence  \cite{bae_if_2022,basu_influence_2020,grosse_studying_2023}.
Specifically, \textbf{ccr} aims to provide a balanced view of influence across content-related and confidence-related training data:
It reduces the bias introduced by documents with extreme (the highest/lowest) influence scores while still capturing variation among the top 10 most relevant documents.  For comparison, we also compute content groundness in the absence of verbalized confidence. This variant, denoted as $ccr^{\neg c}$, is defined analogously using influence estimates obtained without the confidence component (see Equation \ref{eq:negvc}):

\begin{equation}
\resizebox{0.7\linewidth}{!}{$
    ccr^{\neg c}=\frac{\sum_{i=0}^n \sum_{d^t\in T_i, d^f\in F_i}\mathbf{1}(\psi^{\neg c}(d^t|c_i)>\psi^{\neg c}(d^f|c_i))}{\sum_{i=0}^n \sum_{d^t\in T_i, d^f\in F_i}\mathbf{1}(\psi^{\neg c}(d^t|c_i)<\psi^{\neg c}(d^f|c_i))}$}.
\end{equation}

\paragraph{Interpretation of Content Groundness Score.} When $ccr$ (or $ccr^{\neg c}$) $ > 1$, the examined LLM grounds more on content-related training data, whereas $ccr$ (or $ccr^{\neg c}$) $ < 1 $ indicates stronger grounding in confidence-related data. We interpret a higher content groundness score as indicating that the model’s confidence is more influenced by lexically relevant content. However, the threshold for determining whether the confidence of an LLM can be considered trustworthy is subjective and often depends on the specific domain or task. For instance, in high-stakes domains such as medicine, one might require a much higher level of reliability and consider a score of 1.2 insufficient.

\begin{table*}[ht]
\centering
\footnotesize
\caption{Result of LLMs' content groundness measured with \textbf{ccr} and \textbf{ccr}$^{\neg c}$  on different datasets.  We also aggregate all the retrieved data from pre-training and post-training to compute ccr scores for Pre+Post.  The values inside parentheses are insignificant results (p>0.05) in the mean influence score difference between the two comparison sets.\\}\label{tab:main_data_table}
\resizebox{\textwidth}{!}{
\begin{tabular}{p{1.65cm}p{1.3cm}*{10}{c@{\hspace{0.2cm}}c@{\hspace{0.4cm}}}c@{\hspace{0.2cm}}l}
\toprule
\textbf{INS} & \textbf{Data} & 
\multicolumn{2}{c}{\textbf{NQ}} & 
\multicolumn{2}{c}{\textbf{SciQ}} & 
\multicolumn{2}{c}{\textbf{TriviaQA}} & 
\multicolumn{2}{c}{\textbf{TruthfulQA}} & 
\multicolumn{2}{c}{\textbf{PopQA}} & 
\multicolumn{2}{c}{\textbf{All}} \\
\cmidrule(r){3-4} \cmidrule(r){5-6} \cmidrule(r){7-8} 
\cmidrule(r){9-10} \cmidrule(r){11-12} \cmidrule(r){13-14}
\textbf{Model} & \textbf{Source}& ccr & ccr$^{\neg c}$ & ccr & ccr$^{\neg c}$ & ccr & ccr$^{\neg c}$ & ccr & ccr$^{\neg c}$ & ccr & ccr$^{\neg c}$ & ccr & ccr$^{\neg c}$ \\
\midrule
\multirow{3}{*}{OLMo2-13B} & Pre  & 0.76 & 0.75 & 0.71 & 0.71 & 0.84 & 0.82 & 0.73 & 0.70 & 0.74 & 0.73 & 0.75 & 0.74 \\
 & Post & 0.78 & 0.73 & 0.85 & 0.85 & 0.85 & 0.79 & 0.80 & 0.70 & 0.74 & 0.71 & 0.80 & 0.75 \\

 & Pre+Post &  0.77 & 0.75 & 0.78 & 0.78 & 0.84 & 0.81 & 0.77 & 0.70 & 0.74 & 0.72 & 0.78 & 0.75 \\ 
\midrule
\multirow{3}{*}{OLMo2-7B} & Pre &  1.08 & 1.07 & 1.01 & (1.02) & 1.17 & 1.22 & 1.43 & 1.54 & 1.02 & 1.06 & 1.12 & 1.15 \\
 &  Post & 1.85 & 1.95 & 1.63 & 1.52 & 1.73 & 1.87 & 1.67 & 1.84 & 1.74 & 1.71 & 1.72 & 1.77 \\

& Pre+Post  & 1.43 & 1.43 & 1.32 & 1.25 & 1.45 & 1.49 & 1.56 & 1.65 & 1.37 & 1.32 & 1.42 & 1.41 \\
\midrule
\multirow{3}{*}{OLMo-7B} & Pre &  1.22 & 1.22 & (0.95) & (0.96) & 1.08 & 1.08 & 1.30 & 1.29 & 1.06 & 1.06 & 1.11 & 1.11 \\
 & Post & 1.28 & 1.30 & 1.17 & 1.18 & 1.29 & 1.34 & 1.25 & 1.33 & 1.15 & 1.18 & 1.22 & 1.26 \\
& Pre+Post  & 1.25 & 1.25 & 1.05 & 1.05 & 1.18 & 1.19 & 1.27 & 1.30 & 1.10 & 1.11 & 1.16 & 1.17 \\ 
\bottomrule
\end{tabular}}
\end{table*}
\begin{figure}[htbp!]
    \centering
    \includegraphics[width=1\linewidth]{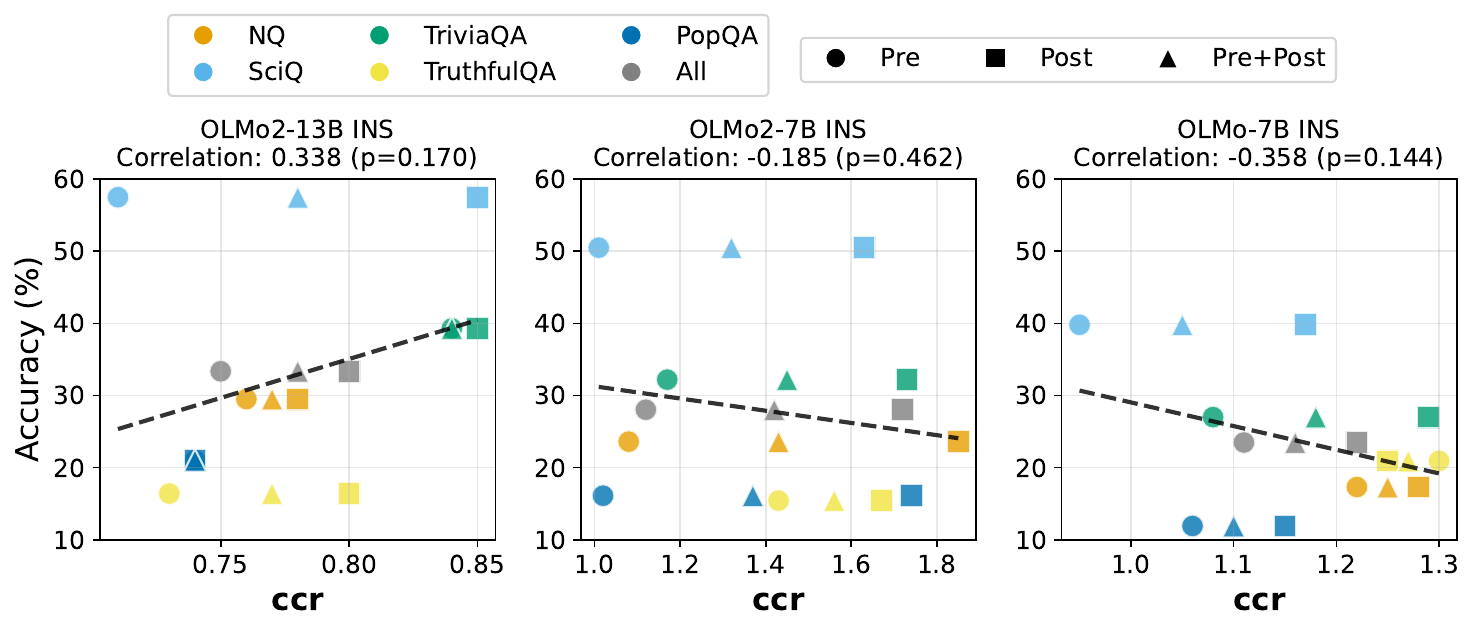}
    \caption{Pearson correlation between content groundness and task accuracy. Each data point reflects the accuracy of the dataset and the ccr score evaluated with this dataset. Each dataset is evaluated with three settings (pre, post and pre+post), thus these three settings can have different ccr scores but the same accuracy.}
    \label{fig:acc_ccr_cor}
\end{figure}

\begin{figure}[ht]
    \centering
    \includegraphics[width=1\linewidth]{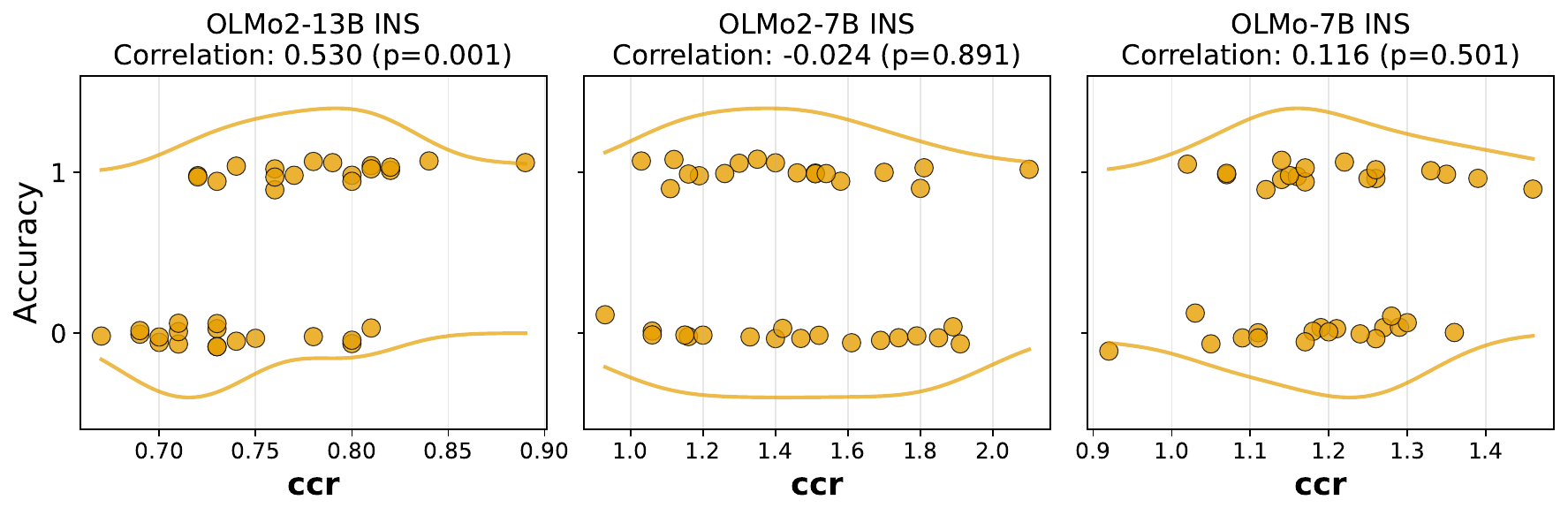}
    \caption{Pearson correlation between content groundness and sample correctness with density estimation plot \cite{scott2015multivariate} (Gaussian kernel). We split the samples of each dataset into a correct set (accuracy = 1) and an incorrect set (accuracy = 0).  Each point presents one set and its corresponding ccr score.}
    \label{fig:correctness_ccr_cor}
\end{figure}

\begin{figure}[ht]
    \centering
    \includegraphics[width=1\linewidth]{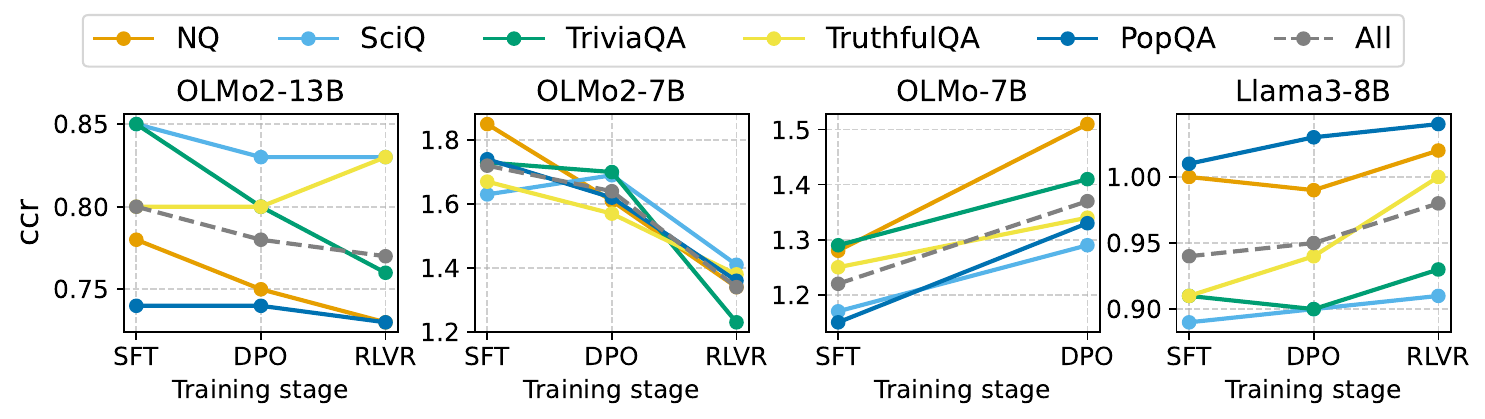}
    \caption{Impact of different post-training schemes on the content groundness. We plot the training stages of the same model over time. The examined training samples are from the post-training data of the corresponding model. 
    We include models of Llama3-8B as their post-training data are publicly available.
    }
    \label{fig:ft-shemes}
\end{figure}
\subsection{What Factors Impact Content Groundness?}
Table \ref{tab:main_data_table} and Fig. \ref{fig:acc_ccr_cor}-\ref{fig:ft-shemes} show the results of different OLMo and Llama models evaluated on five benchmark datasets. We observe from Table \ref{tab:main_data_table} that the differences between \textbf{ccr} and \textbf{ccr}$^{\neg c}$ scores are small, suggesting that \textbf{our method is robust
to variations in the model's self-reported confidence}. Below is a more detailed analysis. 

\paragraph{Larger Models Are Not More Content-grounded.} 

As shown by Table \ref{tab:main_data_table}, confidence generation in OLMo2-13B INS is more influenced by confidence-related training data (ccr < 1), while content-related data plays a greater role in smaller-sized models OLMo-7B and OLMo2-7B INS models (ccr > 1). One possible explanation is that larger models, due to their higher capacity, are more sensitive to stylistic or superficial patterns in training data, such as linguistic markers of certainty, while smaller models, being less expressive, rely more directly on content-aligned signals.

\paragraph{Content Groundness Is Not Correlated with Task Accuracy, but can be impacted by sample correctness.}\footnote{Following previous work \cite{chen-etal-2024-copybench,liu2024litcablightweightlanguagemodel,tian2023just,xia-etal-2025-influences}, we use LLM-as-a-Judge, Prometheus-8x7b-v2.0 \cite{kim2024prometheus}, to assess the semantical equivalence of the generated answers and ground truth answers.} 
As shown in Fig.~\ref{fig:acc_ccr_cor}, the correlations between content groundness and task accuracy are only moderate (Pearson’s $r>0.3$) \cite{datatab2025pearson} and not statistically significant ($p>0.05$). This suggests that content groundness is not correlated with task accuracy. However, we observe a significantly high correlation between content groundness and sample correctness on OLMo2-13B INS from Fig. \ref{fig:correctness_ccr_cor}. Even though this observation does not generalize to other smaller OLMo models.
We hypothesize that when LLMs answer correctly, they may be drawing upon training examples that are similar to the input, leading them to ground their expressed confidence more heavily in content-relevant information.

\paragraph{Post-training Schemes Can Impact Content Groundness.} 

Fig. \ref{fig:ft-shemes} demonstrates that the impact of post-training schemes on content groundness can be opposite on different models. For example, DPO or RLVR improves the content groundness for OLMo-7B and Llama3-8B but not for OLMo2-13B and OLMo2-7B models. We also observe that post-training schemes generally do not reverse the content groundness results, i.e., scores are either all under one or above one, which may indicate that pre-training schemes are more likely to determine if content groundness is greater than 1.

\section{Discussions}
This section offers a discussion aimed at deepening the understanding of our method and addressing the limitations.

\paragraph{Practical Constraints.} Our findings are constrained to a limited set of LLMs, primarily the OLMo and Llama families, due to restricted access to the pre-training and post-training data of other proprietary models. Since TracVC relies on analyzing training data influence, the lack of transparency and availability of training data for widely used commercial LLMs (e.g., GPT-4, Claude, Gemini) limits the generalizability of our conclusions. Note that this is not a weakness of our method but rather a practical constraint.  Future work could extend this analysis as more open-source models and datasets become available.

\paragraph{Simplified Separation of Content- and Confidence-Related Influences.} TracVC approximates content- and confidence-related influences using two disjoint sets of retrieved documents. In practice, these influences may overlap, and a single training example may encode both content- and confidence-related signals. Our decomposition should therefore be viewed as a coarse-grained approximation rather than a strict separation of internal model mechanisms. 

\paragraph{Limited Coverage of Studied Samples and Influence Estimation Methods.} Our retrieval stage relies on BM25-based lexical matching, which emphasizes surface-level word overlap. As a result, semantically relevant but lexically dissimilar training examples may be omitted, while examples with shared keywords but limited relevance may be retrieved. Consequently, our influence estimates are confined to a retrieval-constrained subset of the training data rather than the full corpus. We choose to retrieve a fixed number of training examples ($k = 10$) per category. Although this choice balances computational efficiency and interpretability, it is inherently heuristic and may lead to variability in the estimates.
Furthermore, we employ a gradient-based influence estimation method adapted from TracIn for scalability. While this approach is well grounded in prior work, we do not compare it against alternative influence estimation techniques, such as Hessian-based methods, which may offer greater precision but do not currently scale to LLMs of the size considered in this study. Overall, \textbf{our method should be viewed as an initial step toward understanding the training-data origins of verbalized confidence, rather than a comprehensive solution.}

\section{Conclusion}
LLMs often exhibit overconfidence, raising the question of whether their confidence is grounded in content-relevant training data. In this paper, we introduced TracVC, a method for tracing the origins of LLM confidence back to specific types of training examples. Our analysis shows that OLMo2-13B models are more influenced by confidence-related than content-related data when estimating confidence, revealing a potential source of miscalibration and risk to trustworthiness. We also find that larger LLMs are not necessarily more content-grounded, and that content groundness can be affected by sample correctness and the extent of post-training.
These findings underscore the need for future research on how pre-training and fine-tuning strategies shape confidence grounding, with the ultimate goal of developing models whose confidence more reliably reflects over content. 

More broadly, TracVC offers a general, data-centric approach for interpreting LLM behavior through the lens of training data influence, and can be extended to study other aspects of model behavior, such as answer selection and reasoning patterns. We view TracVC as an initial step toward training-data-level explanations for more transparent and trustworthy LLMs.

\section*{Appendix}
\paragraph{Details of the LLM Training Data.}

Table \ref{tab:search_data} summarizes the pre-training and post-training datasets used in our analysis. Given the massive scale of the pre-training corpus, we do not construct our own index but instead rely on the publicly available inverted index provided by Elazar et al. \cite{DBLP:conf/iclr/ElazarBMRSSWGS024}, which was built over the Dolma v1.7 dataset.

\begin{table}

     \caption{Details of the pre-training and post-training data used for retrieving data examples in the paper.\\}\label{tab:search_data}
         \centering
    \scriptsize
    \resizebox{\textwidth}{!}{\begin{tabular}{ccp{8cm}}
    \toprule
     Model    & Pre-training Data & Post-training Data\\
     \midrule
    OLMo2-13B INS      & dolma v1.7   &tulu-3-sft-olmo-2-mixture, olmo-2-1124-13b-preference-mix, RLVR-MATH\\
    OLMo2-13B-DPO     &  dolma v1.7 & tulu-3-sft-olmo-2-mixture, olmo-2-1124-13b-preference-mix\\
    OLMo2-13B-SFT     &   dolma v1.7& tulu-3-sft-olmo-2-mixture\\
    Llama3-8B-INS     &  -& tulu-3-sft-mixture, RLVR-GSM-MATH-IF-Mixed-Constraints\\
    Llama3-8B-DPO     &  -&tulu-3-sft-mixture, llama-3.1-tulu-3-8b-preference-mixture, llama-3.1-tulu-3-8b-preference-mixture \\
    Llama3-8B-SFT     & - & tulu-3-sft-mixture\\
    OLMo2-7B-INS     &  dolma v1.7 & tulu-3-sft-olmo-2-mixture, olmo-2-1124-7b-preference-mix, RLVR-GSM\\
    OLMo2-7B-DPO     & dolma v1.7  & tulu-3-sft-olmo-2-mixture, olmo-2-1124-7b-preference-mix\\
    OLMo2-7B-SFT     &  dolma v1.7 &tulu-3-sft-olmo-2-mixture \\
    OLMo-7B-INS     & dolma v1.7 & tulu-v2-sft-mixture, ultrafeedback-binarized-cleaned\\
    OLMo-7B-SFT     & dolma v1.7& tulu-v2-sft-mixture\\
    \bottomrule
    \end{tabular}}
   
\end{table}

\begin{credits}
\subsubsection{\ackname} This research has been funded by the Vienna Science and Technology Fund (WWTF) [10.47379/VRG19008] ``Knowledge infused Deep Learning for Natural Language Processing''.
\end{credits}
\clearpage

\bibliographystyle{splncs04}

\bibliography{custom_de_duplicated_sub}
\end{document}